\documentclass{article} 
\usepackage{iclr2021_conference,times}

\iclrfinalcopy

\usepackage{amsmath,amsfonts,bm}

\def\eqref#1{equation~\ref{#1}}

\def\1{\bm{1}}

\DeclareMathAlphabet{\mathsfit}{\encodingdefault}{\sfdefault}{m}{sl}
\SetMathAlphabet{\mathsfit}{bold}{\encodingdefault}{\sfdefault}{bx}{n}

\DeclareMathOperator*{\argmin}{arg\,min}

\usepackage{multirow}
\usepackage{todonotes}
\usepackage{hyperref}
\usepackage{url}
\usepackage{mathtools}
\usepackage{amssymb}
\usepackage{optidef}
\usepackage{multicol}
\usepackage{booktabs}
\usepackage{nicefrac}
\usepackage{graphicx}
\usepackage{todonotes}
\usepackage[linesnumbered,ruled,vlined]{algorithm2e}
\usepackage{algorithmic}
\usepackage{cleveref}
\SetKw{KwBy}{by}

\title{
Making Differentiable Architecture Search less local
}

\author{Erik Bodin \thanks{This work was done during an internship at Spotify.} \\
Spotify \& University of Bristol \\
\And
Federico Tomasi \\
Spotify \\
\And
Zhenwen Dai \\
Spotify \\
}

\begin{document}

\maketitle

\begin{abstract}
Neural architecture search (NAS) is a recent methodology for automating the design of neural network architectures.
Differentiable neural architecture search (DARTS) is a promising NAS approach that dramatically increases search efficiency.
However, it has been shown to suffer from performance collapse, where the search often leads to detrimental architectures.
Many recent works try to address this issue of DARTS by identifying indicators for early stopping, regularising the search objective to reduce the dominance of some operations, or changing the parameterisation of the search problem.
In this work, we hypothesise that performance collapses can arise from poor local optima around typical initial architectures and weights.
We address this issue by developing a more global optimisation scheme that is able to better explore the space without changing the DARTS problem formulation.
Our experiments show that our changes in the search algorithm allow the discovery of architectures with both better test performance and fewer parameters.
\end{abstract}

\section{Introduction}
\label{sec:introduction}

Designing neural network architectures improving upon the state-of-the-art requires a substantial effort of human experts.
Automating the discovery of neural network architectures by formulating it as a search problem allows us to minimise the human time spent on the search process. 
Due to the large combinatorial search space of possible neural network architectures, early methods~\cite{zoph2016neural,zoph2018learning,real2019regularized} were computationally very demanding, 
often requiring thousands of GPU days of computation for search, giving rise to high costs.
Many neural architecture search (NAS) works have been focused on reducing the computational cost,~\cite{liu2018progressive,bender2018understanding,elsken2017simple,pham2018efficient,cai2017efficient}.
Among them, Liu et al. \cite{liu2018darts} proposed a particularly efficient approach by making the search space of architectures differentiable (known as DARTS), which reduced
the search cost by several orders of magnitude.

Although being efficient, recent works have shown that DARTS suffers from performance collapse due to the search favouring parameter-less operations like skip connections \cite{chu2020fair,zela2019understanding}. Many follow-up works have been proposed to fix the performance collapse problem by identifying indicators for early stopping, regularising the search objective to reduce skip connections, or changing the search problem's parameterisation.
Chen \& Hsieh~\cite{pmlr-v119-chen20f} and Zela et al.~\cite{zela2019understanding}  proposed to stabilise the search process by regularising the Hessian of the search objective. Chu et al.~\cite{chu2020fair} avoid the advantage of the skip connections in the search phrase by replacing the softmax with the sigmoid function for the switch among edges.
Chu et al.~\cite{chu2021darts} avoided the dominance of skip connections by changing the parameterisation of the search space.

In this paper, we hypothesise that performance collapses and the dominance of some operations observed in several works are the consequence of the existence of poor local optima around typical initial architectures and weights. Instead of identifying indicators for early stopping or tweaking the search space's parameterisation, we propose that a more global optimisation scheme should be developed that allows us to avoid bad local optima and better explore the objective over the search space to discover better solutions. 
We show in experiments that even a simple scheme to make the optimisation more global reduces detrimental behaviours significantly.
Importantly, 
it removes the need to stop the search early in order to avoid reaching detrimental or invalid solutions.
We show that, after searching until convergence, our method can find architectures with better test performance and fewer parameters. 

\section{Empirical diagnosis}
\label{sec:empiriacl_diagnosis}

\begin{figure}[t]
\vspace{-5mm}
    \centering
    \includegraphics[trim=5 5 5 5, clip, width=0.3\textwidth]{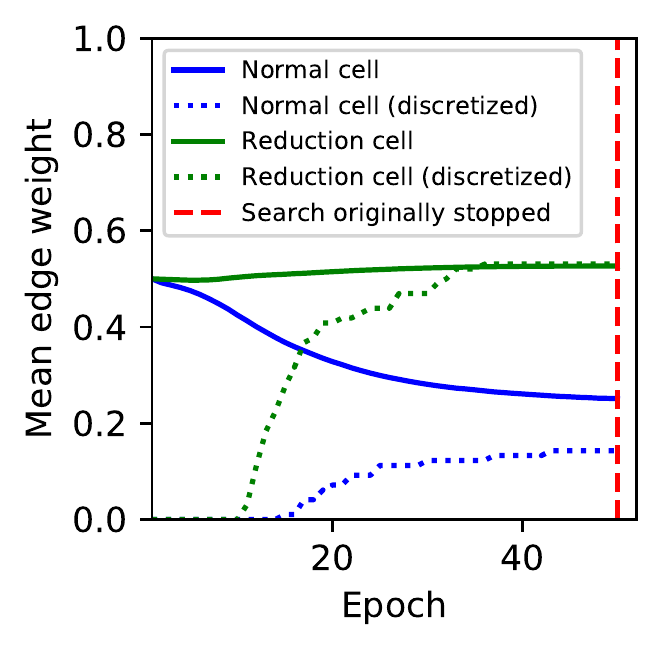}
    \includegraphics[trim=5 5 5 5, clip, width=0.3\textwidth]{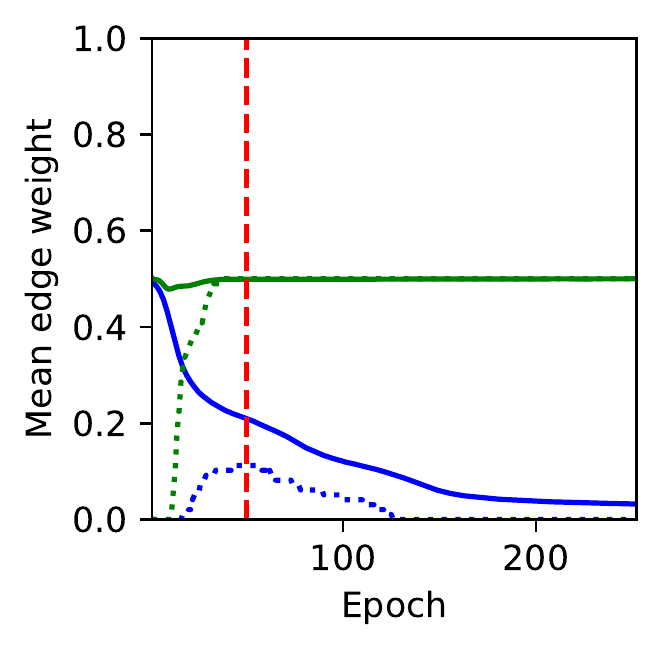}
    \includegraphics[trim=5 5 5 5, clip, width=0.3\textwidth]{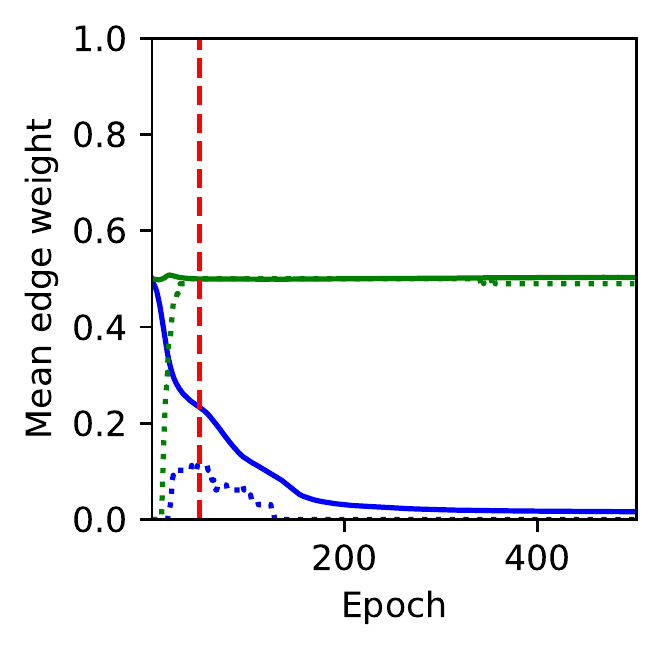}
    \caption{
    The FairDARTS search needs to be stopped early to avoid the normal
    cell having no remaining active edges following discretisation.
    Shown is the mean cell weights (following the sigmoid activation function) for the normal and reduction cell, respectively, for three runs on different budgets.
    The same issue persisted on every run.
    The discretisation threshold used is 0.85, but the issue applies to any thresholding rule as all normal cell edge weights tend to zero.
    }
    \label{fig:fairdarts_sinking_ship_new}
\end{figure}

FairDARTS is a state-of-the-art DARTS variant presented in~\cite{chu2020fair}.
The method includes structural changes to the original DARTS search space, 
allowing multiple edges per pair of nodes in the searched cell structure.
This was implemented by switching the softmax activation function on the weights to a sigmoid function.
Another change was adding a regularisation term (a `zero-one loss'), 
encouraging the continuous edge activations to better approximate the binary discretisation of the cell happening after the search phase.

FairDARTS improved upon the DARTS method, reducing the issue with skip connections dominating, 
and ultimately lead to better architectures in terms of final test performance compared to DARTS and other variants.
However, as we will demonstrate, another similar issue presents itself (still) in the FairDARTS method.
What happens is that one of the searched cell types, the `reduction cell', dominates the other (the `normal cell'), to the detriment of test performance and reliability.
In particular, if searching for longer than a small fraction of as many epochs later used for the training in the final evaluation phase,
the test performance decays, and the architectures produced quickly become invalid.
We illustrate this in \Cref{fig:fairdarts_sinking_ship_new} (the experiments were conducted on CIFAR-10 using the implementation and the setup as in~\cite{chu2020fair}).
We note that the edge weights associated with the normal cell decrease monotonically after a certain number of epochs.
If the search is not stopped early, at the right time, the weights of all operations in the normal cell become zero, 
resulting in no activations being able to propagate through the cell following discretisation. 
In~\cite{chu2020fair} the search was stopped after only $1 / 12$ of the number of epochs later used to train the final architecture.

Being forced to stop the search early to avoid detrimental architectures has two negative consequences.
Firstly, the right time to stop the search becomes an additional hyperparameter to tune to obtain good performance.
Secondly, it can inhibit better architectures to be found by searching for longer.
Both of these aspects are important for building a reliable NAS method for a wide range of datasets and tasks.

\section{Global Optimisation for Differentiable NAS}

DARTS~\cite{liu2018darts}, 
as similar to prior works~\cite{zoph2018learning,real2019regularized,liu2018progressive},
searches for a \emph{cell}, which is used as a building block for the final architecture.
The cell constitutes a directed acyclic graph of $N$ nodes.
Each node $x$ represents a latent representation
and each directed edge $(i, j)$ represents an operation $o_{i, j}$.
A node depends on all of its predecessors as $x_j = \sum_{i < j} o_{i, j}(x_i)$.
Let $\mathcal{O}$ be the set of candidate operations (e.g., convolution, max pooling, skip connection) available for each edge $(i, j)$.
FairDARTS~\cite{chu2020fair} defines the choice of operations for an edge as
$\bar{o}_{i, j}(x) = \sum_{o \in \mathcal{O}} \sigma(\alpha_{o_i,j})o(x)$,
where $\sigma(\cdot)$ is the sigmoid function. 
This allows multiple operations per edge to be chosen simultaneously.
If no operations are active for a given edge this constitute the zero operation~\cite{zela2019understanding}.

Let $\bm{\alpha}$ be the concatenated vector of all operation edge weights representing the architecture, 
in which the ones associated with the normal cell and the reduction cell are denoted by $\bm{\alpha}_{\text{normal}}$ and  $\bm{\alpha}_{\text{reduction}}$ respectively, i.e., $\bm{\alpha} = (\bm{\alpha}_{\text{normal}}, \bm{\alpha}_{\text{reduction}})$.
Let $\bm{w}$ be the concatenated neural network parameters associated with all operations, where similarly $\bm{w} = (\bm{w}_{\text{normal}}, \bm{w}_{\text{reduction}})$.

The architecture search problem in DARTS can be stated as a bilevel optimisation problem:
\begin{mini!}|l|[2]
{\bm{\alpha}}
{
\mathcal{L}_{\text{val}}(\bm{\alpha}, \bm{w}^{\ast})
}
{}{
\label{eq:bilevel_1}
}
\addConstraint{
\bm{w}^{\ast} = \argmin_{\bm{w}} \mathcal{L}_{\text{train}}(\bm{\alpha}, \bm{w})
},
\label{eq:bilevel_2}
\end{mini!}
where $\mathcal{L}_{\text{val}}$ and $\mathcal{L}_{\text{train}}$ are the validation loss and training loss, respectively.
DARTS approximates the gradient as
$
    \nabla_{\bm{\alpha}} \mathcal{L}_{\text{val}}(\bm{\alpha}, \bm{w}^{\ast}) 
    \approx 
    \nabla_{\bm{\alpha}} \mathcal{L}_{\text{val}}(
    \bm{\alpha}, 
    \bm{w} - 
    \xi \nabla_{\bm{w}} \mathcal{L}_{\text{train}}(\bm{\alpha}, \bm{w})
    )
$, 
where $\xi$ is the learning rate for the inner optimisation,
and gradient-based local optimisation is performed in alternating steps.

\paragraph{Global Optimisation Scheme.}
\label{sec:globalopt}
We hypothesise that the usage of local search for the $\bm{\alpha}$ weights in the DARTS' approximation to the bilevel optimisation problem leads to convergence to local optima associated with performance collapse.
We propose an optimisation scheme that makes the search for the $\bm{\alpha}$ weights ``more global" in the sense that local valleys can be escaped using a complementary global optimisation routine.

\begin{algorithm}[t]
  \caption{Doubly Stochastic Coordinate Descent (global step)}
  \label{algo:doubly_coord}
  \SetAlgoLined
  \SetNoFillComment
  \KwIn{Function $f$ defined over $\mathcal{X}$, proposal distribution $q$, initial $\bm{x}_{\text{best}}, y_{\text{best}}$}
  \KwOut{$\bm{x}_{\text{best}}, y_{\text{best}}$}
  \While{budget\_remaining}{
  $d$ = sample\_a\_random\_dimension()\;
  $\bm{x} \sim q(x|\bm{x}_{\text{best}}[d], d)$\;
  $y = f(\bm{x})$\;
  \If{$y < y_{\text{best}}$}{
    $\bm{x}_{\text{best}} = \bm{x}$, $y_{\text{best}} = y$\;
  }
  }
\end{algorithm}

Our optimisation scheme consists of two types of steps: \emph{local} and \emph{global} steps. 
The algorithm alternates between taking local and global steps, 
similar to basin-hopping~\cite{wales1997global} for global optimisation.
A local step is a step in the gradient direction, the same as in~\cite{liu2018darts}.
A global step is taken according to the proposed doubly stochastic coordinate descent (DSCD) algorithm. 
DSCD follows the stochastic coordinate descent approach~\cite{nesterov2012efficiency} and draws a random dimension of which to consider next.
In DSCD, only a single (global) step is taken each time a dimension is sampled, 
and the step is stochastic, 
where the new position (for the sampled dimension) is a sample from a proposal distribution.
The sample is accepted as the new position only if the objective improves upon the best lost within the last K steps
\footnote{In practice we used $K = 1000$. 
Only considering the best loss within a (relatively large) window, rather than the historical best, we noted was helpful to be robust to outlier losses as a result of the mini-batching.
}. 
The global step is global in the sense that there does not need to be a monotonically improving trajectory between any two positions (in terms of the loss surface), thus allowing `jumps' between valleys
\footnote{Strictly this does not need to be true for \emph{stochastic gradient descent} either, but in SGD it is still statistically unlikely to take steps in non-monotonically improving directions.}
.
The outline of DSCD is shown in Algorithm~\ref{algo:doubly_coord}.
We propose an annealing scheme for the proposal distribution. The proposal distribution is parameterised as a Beta distribution over a bounded space. At the beginning of the optimisation, the proposal distribution is uniform, and it slowly moves towards a Dirac delta centred at the current position, thus becoming increasingly local as the search progresses. The details of the annealing scheme for the proposal can be found in the appendix.
We alternate between taking local and global steps when the following is both true; $T$ consecutive steps of the same type has been taken, and the loss did not improve from the last step to the next.
In all experiments, we set $T = 50$, and noticed little to no importance of tuning this parameter.
In the appendix we assess the benefit of DSCD on multimodal functions.

\section{Experiments}

\begin{figure}[t]
\vspace{-10mm}
    \centering
    \includegraphics[trim=5 5 5 5, clip, width=0.3\textwidth]{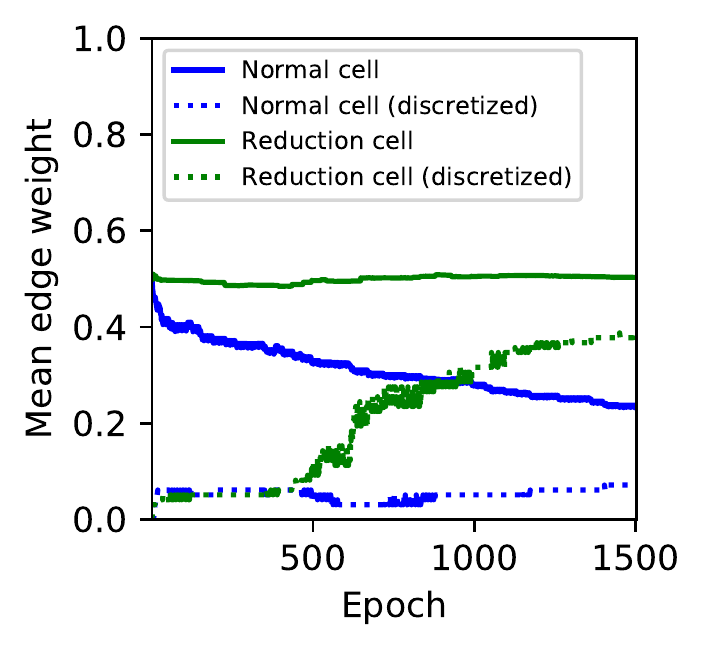}
    \includegraphics[trim=5 5 5 5, clip, width=0.3\textwidth]{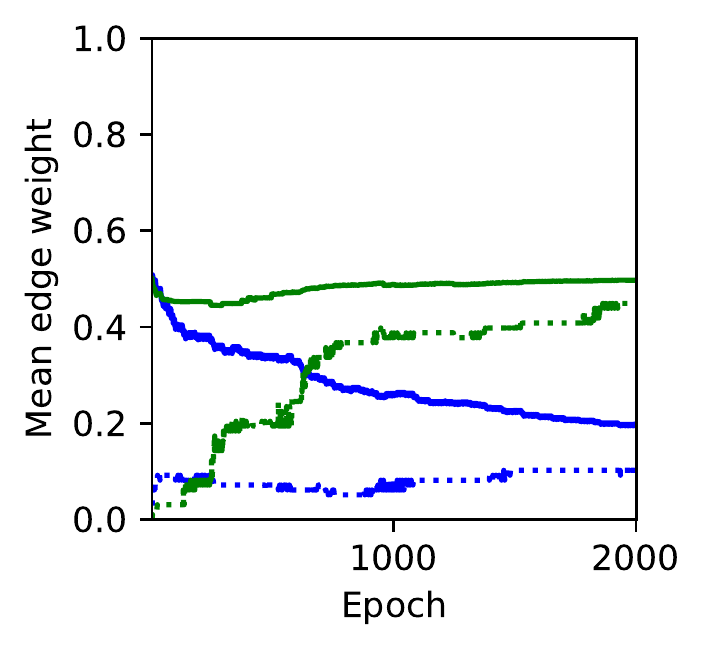}
    \includegraphics[trim=5 5 5 5, clip, width=0.3\textwidth]{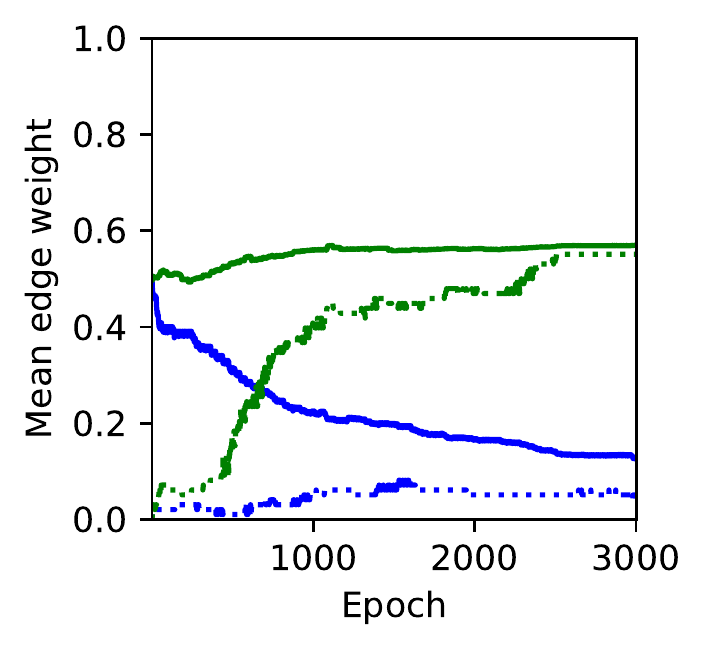}
    \caption{
    Using the new optimisation scheme the architecture does no longer become invalid by searching for longer.
    Shown is the mean cell weights (following the sigmoid activation function) for the normal and reduction cell, respectively, on three runs on different budgets.
    }
    \label{fig:fairdarts_using_our_search}
\end{figure}

We previously showed that all the edge weights of normal cells $\bm{\alpha}_{\text{normal}}$ tend towards zero in FairDARTS, resulting in invalid architectures.
We will now demonstrate that our optimisation scheme explores the architecture space better. 
As a result, it avoids invalid architectures, discovers architectures with better test performance, and converges to good solutions without early stopping.
In the experiments, the same setup as in~\cite{chu2020fair} is used, except for ``FairDARTS + DSCD" for which we replace the local optimiser (Adam~\cite{KingmaB14}) with the proposed optimisation scheme.
In \Cref{fig:fairdarts_using_our_search} we see that the edge weights of the normal cell no longer become zero, even if searching for much longer,
and the resulting architecture can be successfully discretised.
The mean weights, after discretisation, slowly move towards the mean weights before discretisation.
Importantly, the edges that will be kept (above the $0.85$ threshold) remained the same from $1500$ epochs,
which is indicative of convergence.

\begin{table}[t]
\vspace{-4mm}
\centering
\caption{
Comparison with FairDARTS for search and evaluation phases (accuracy in \%).
Split for $\mathcal{L}_{\text{train}}$ and $\mathcal{L}_{\text{val}}$ indicates accuracy measured on the training data for $\mathcal{L}_{\text{train}}$ and $\mathcal{L}_{\text{val}}$ respectively. Search Test indicates the accuracy on the hold-out set using the search network (undiscretised). Eval. Test indicates the test accuracy with the final architecture.
``Invalid arch." denotes no valid final architecture after discretisation.
}
\label{table:results_acc}
\vspace{0.05cm}
\resizebox{0.85\linewidth}{!}{
\begin{tabular}{ c | c c c | c }
\toprule
\multirow{2}{*}{Method}  & \multicolumn{3}{|c|}{Search Phase} & Final Arch.  \\
&Split for $\mathcal{L}_{\text{train}}$ & Split for $\mathcal{L}_{\text{val}}$ & Search Test & Eval. Test\\ \midrule
 FairDARTS (50) & 82.02 & 75.61 & 76.15 & 97.36 \\ 
 FairDARTS (75) & 87.35 & 78.10 & 78.65 & 97.29 \\
 FairDARTS (250) & 96.95 & 81.55 & 81.52 & Invalid arch. \\
 FairDARTS (500) & 99.92 & 83.49 & 83.26 & Invalid arch. \\
 FairDARTS + DSCD (1500) & 100.0 & 83.12 & 83.40 & 97.50 \\
 FairDARTS + DSCD (2000) & 100.0 & 84.02 & 84.71 & 97.25 \\
 FairDARTS + DSCD (3000) & 100.0 & 85.51 & 85.10 & 96.92 \\
 \bottomrule
\end{tabular}
}
\end{table}

In \Cref{table:results_acc} we see the accuracy of the final architectures and the searches, corresponding to \Cref{fig:fairdarts_sinking_ship_new,fig:fairdarts_using_our_search}.
Using our optimisation scheme (DSCD), the models produced become increasingly more accurate with more search, 
while remaining valid.
Our method using 1500 epochs for search produces a higher test accuracy during the search phase than FairDARTS, which also results in a high test accuracy with the final architecture.
Despite the test accuracy of our method increasing with more search epochs, the test accuracy of the resulting final architectures decreases.
We argue that this is due to the fact that 
the network used during the search phase is different from the network for evaluation (a network trained from scratch using the final architecture)~\cite{chu2020fair}.
Differences between the search architecture and final architecture include discretisation, 
that the final architecture is larger and has auxiliary heads~\cite{chu2020fair},
as well as that the training paths are different (weights and architecture together versus weights only).
A comparison to other DARTS variants is included in the appendix.

\section{Conclusion}

Neural architecture search requires three things: 
a space of models with good inductive biases, 
a loss function to assess models, 
and an optimisation or inference algorithm to explore the space.
In this work we focused on the optimisation algorithm, and we showed that by combining gradient-based, 
local search with global optimisation techniques, 
we are able to better explore the space.

\bibliographystyle{abbrv}
\bibliography{main}

\appendix

\begin{center}
{\Large Appendix}
\end{center}

\section{Comparison with other DARTS methods}

We also compared our approach with other state-of-the-art NAS methods in the DARTS family. The results are shown in Table~\ref{table:results_comp}.

\begin{table}[h]
\centering
\caption{
Comparison of state-of-the-art NAS models on CIFAR-10.
FairDARTS$\ast$ differs from FairDARTS in that the former uses additional post-processing of the edge weights after search,
with a hard limit on the number of edges kept per node pair.
}
\label{table:results_comp}
\vspace{0.2cm}
\begin{tabular}{ c c c c }
 Method & Params (M) & FLOPS (M) & Accuracy (\%) \\ \hline
 DARTS~\cite{liu2018darts} & 3.3 & 528 & 97.00 \\
 DARTS-~\cite{chu2021darts} & 3.5 & 583 & 97.41 \\
 FairDARTS$\ast$~\cite{chu2020fair} & 2.8 & 373 & 97.46 \\ 
 FairDARTS & 6.4 & 966 & 97.36 \\ 
 FairDARTS + DSCD & 3.6 & 532 & 97.50
\end{tabular}
\end{table}

\section{Assessment of DSCD on multimodal functions}
\label{sec:known_funcs}

\begin{figure}[t]
\vspace{-10mm}
    \centering
    \includegraphics[width=0.99\textwidth]{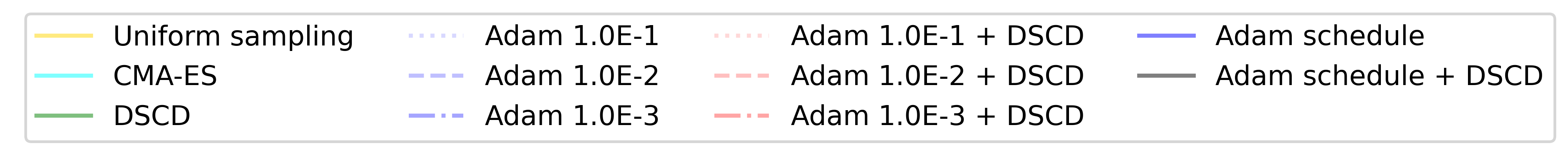}\\
    \includegraphics[trim=8 5 8 5, clip, width=0.48\textwidth]{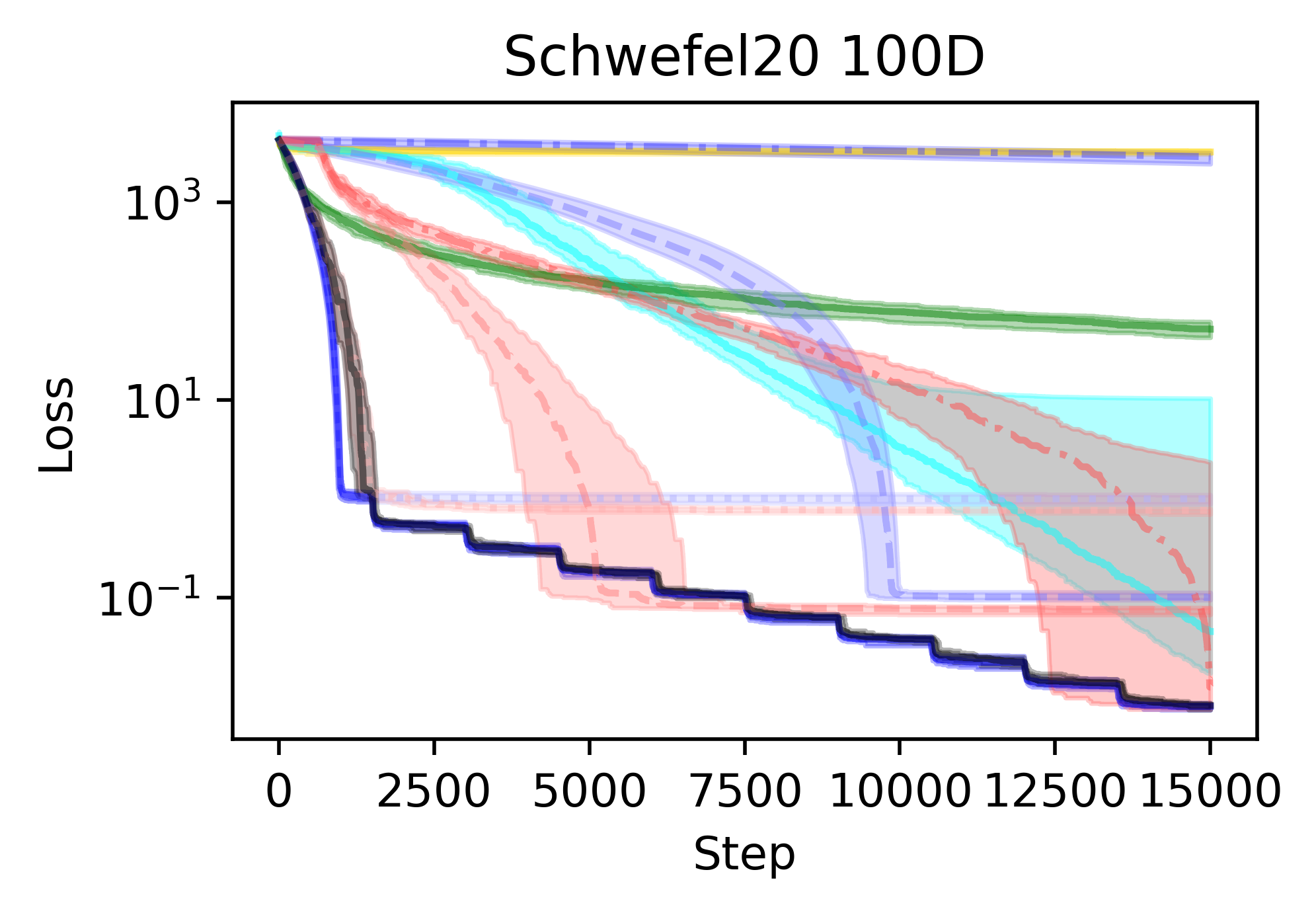}
    \includegraphics[trim=8 5 8 5, clip, width=0.48\textwidth]{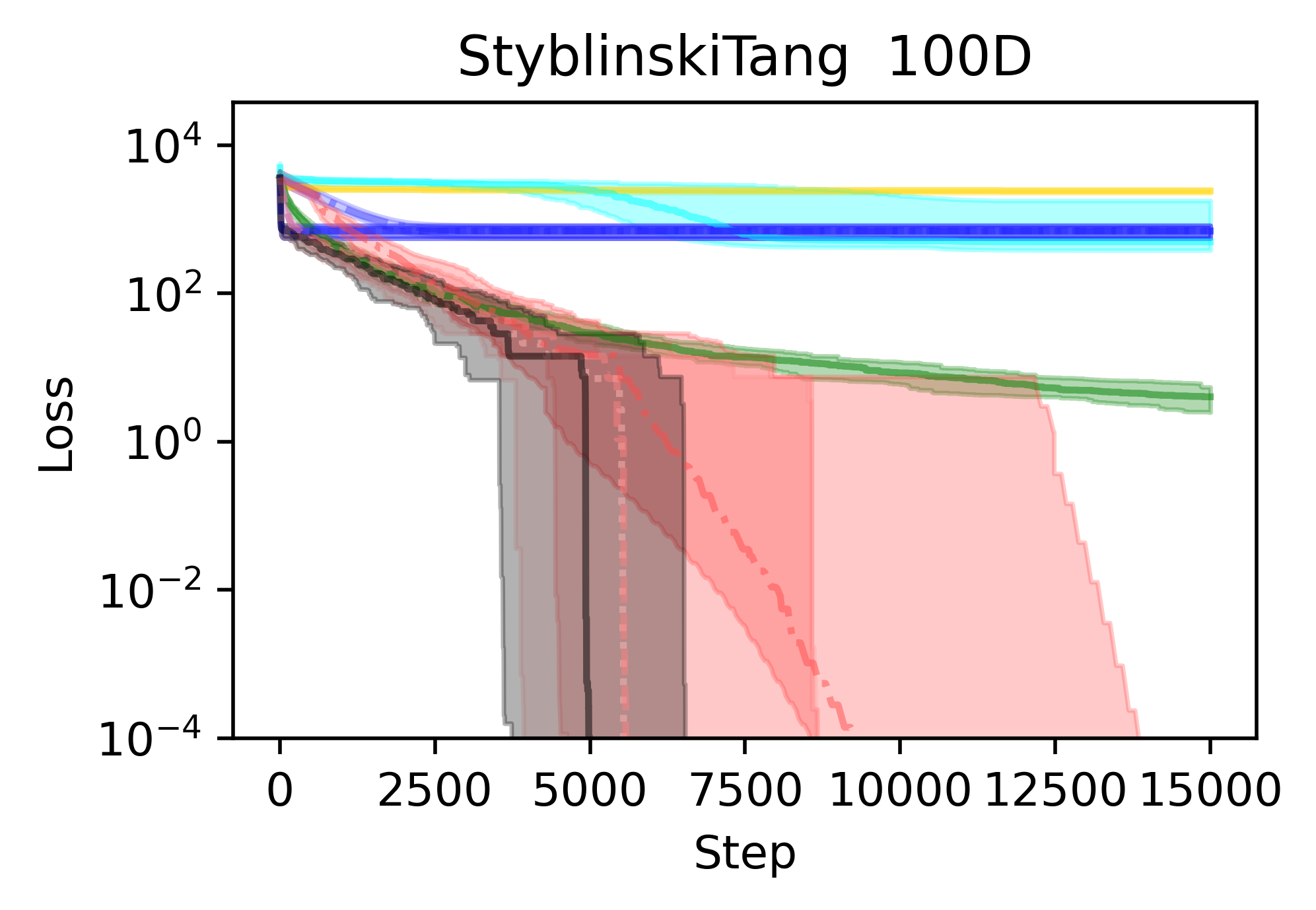}\\
    \includegraphics[trim=8 5 8 5, clip, width=0.48\textwidth]{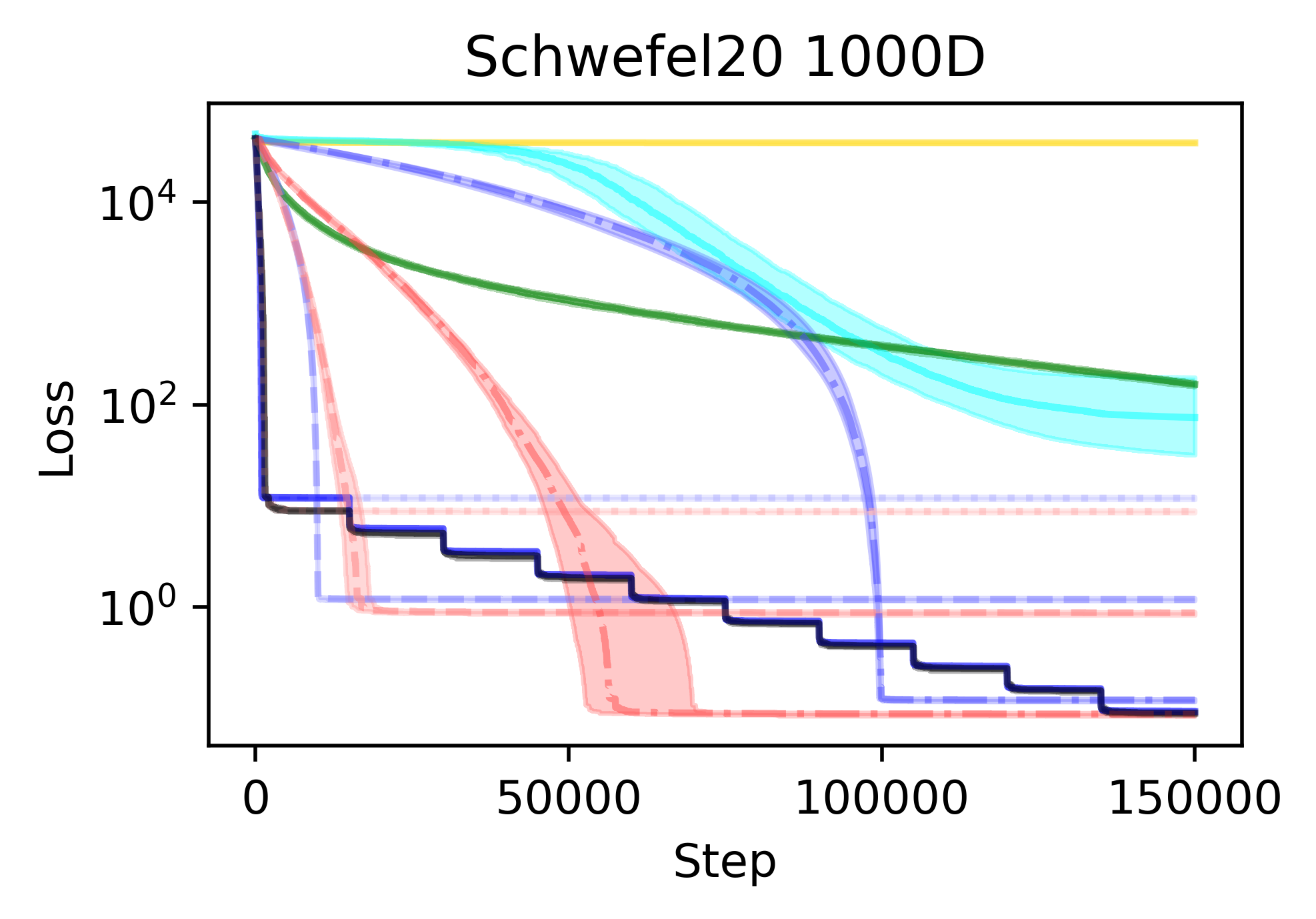}
    \includegraphics[trim=8 5 8 5, clip, width=0.48\textwidth]{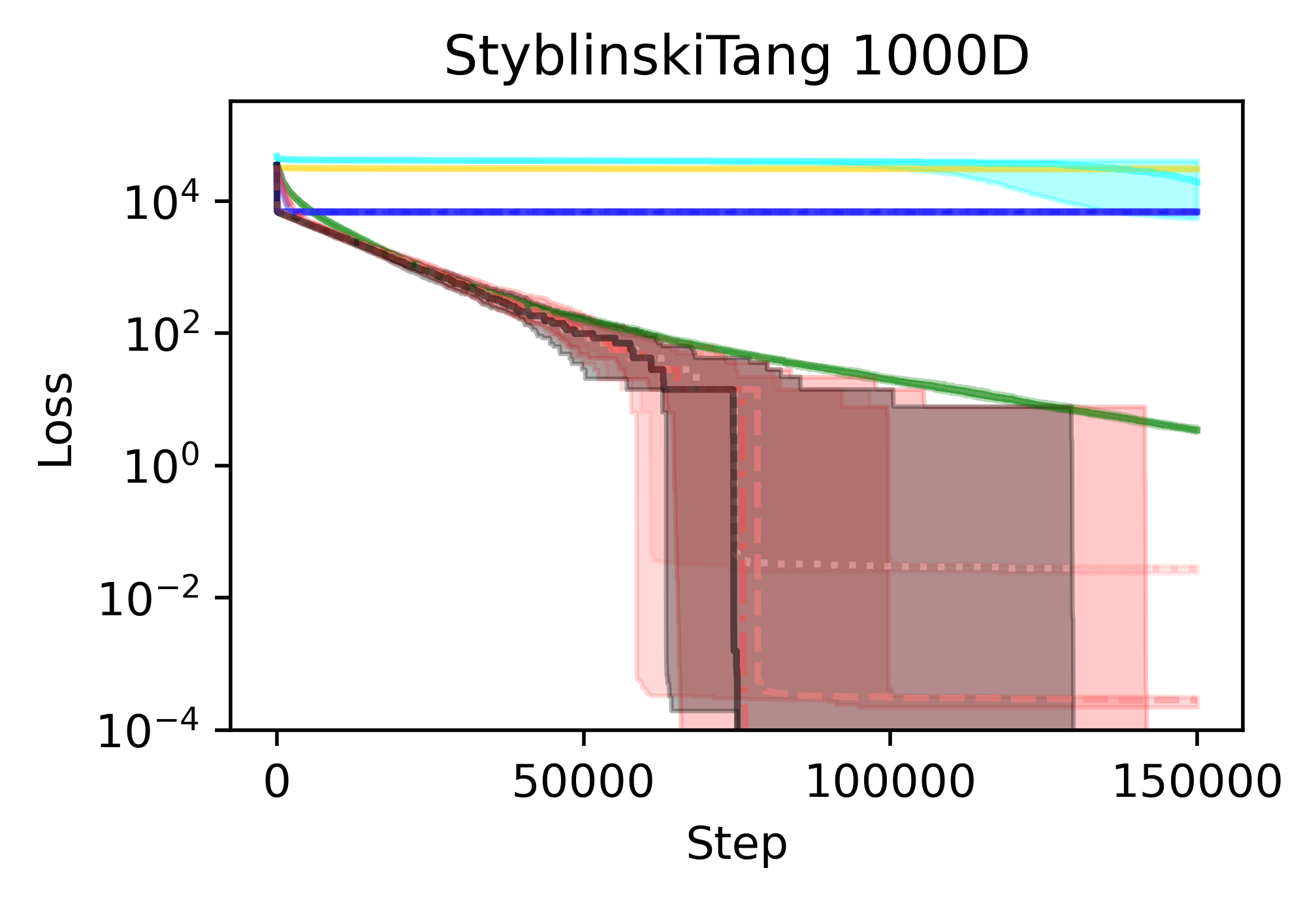}
    \caption{
    Shown is the median loss of 20 runs from uniformly sampled initial positions. 
    Shaded areas display the 95\% CI of the median. 
    The numbers following ``Adam" for each entry in the legend denote the used learning rate,
    where ``schedule" denotes a linear learning rate scheduling between $0.001$ and $0.1$.
    The postfix ``+ DSCD" denotes complementing the method with DSCD (Section~\ref{sec:globalopt}).
    }
    \label{fig:synth_1}
\end{figure}

To confirm and quantify the beneficial effect of complementing gradient-based, local optimisation (Adam) with the proposed doubly stochastic coordinate descent (DSCD) routine, we performed comparisons with and without the routine on synthetic functions with known properties.
For reference, we compare to performing uniform sampling over the domain, as well as Covariance Matrix Adaptation Evolution Strategy (CMA-ES)~\cite{hansen2004evaluating},
a popular global optimisation method.

In Figure~\ref{fig:synth_1} we show the results on the Styblinski-Tang function~\cite{styblinski1990experiments} and the Schwefel function~\cite{jamil2013literature},
which are popular functions for benchmarking optimization methods. 
Both functions have several local minima that are worse than the global minimum.
We note that for every setting of Adam with a particular learning rate,
or using a learning rate schedule, 
complementing the local steps with DSCD global steps (Section~\ref{sec:globalopt}) improves the performance.
On the Styblinski-Tang function, the difference is dramatic, as all the Adam variants without DSCD become stuck in a bad local minimum at every run.

\section{Beta annealing}
\label{sec:beta}

For setting the proposal distribution, we propose an annealing scheme, which we will refer to as \emph{Beta annealing}.
The idea is that, at each step, we will sample a new (scaled) position following a Beta distribution, parameterised to have a varied concentration around the current position.

In practice, for our specific problem of setting operation edge weights going through a sigmoid, we set the proposal domain as $[-3, 3]$ for every dimension,
which accounts for the region of the domain with a significant effect on the output.
Note, however, that positions outside the domain are still possible to reach as of the local optimiser, although position outside will not be proposed in this step.

The current position we (min-max) normalize using the domain, so that it corresponds to a unit position $\upsilon_{i} \in [0, 1]$.
The new proposal unit position, which we address below, is then mapped back to the original domain before the loss evaluation.

We define a concentration parameter $\phi \in [0, 1)$, 
where $\phi = 0$ correspond to an uniform distribution of the (unit) domain,  
and $\phi \rightarrow 1$ tends towards a Dirac delta located at the current position.
The former represents full global exploration (of the sampled dimension), independent of the current position.
The latter represents full local exploitation at the current position.
These two extremes are represented as parameterisations of a Beta distribution,
and all the intermediate settings are as well.
During search we start with $\phi = 0$ and anneal towards $\phi = 1$ at the final epoch.
The annealing schedule used for $\phi$ is cosine annealing, typically used for learning rate scheduling~\cite{loshchilov2016sgdr}.

The proposal (unit) position is sampled as $\upsilon_{i + 1} \sim Beta(\alpha_i, \beta_i)$,
where the $\alpha_i, \beta_i$ parameters depend on $\phi$ and the current (unit) position $\upsilon_{i}$.
Specifically, $\alpha_i, \beta_i$ is derived at each step as following.

The two extremes, the uniform ($\phi = 0$) and Dirac delta ($\phi = 1$),
have known $\alpha$ and $\beta$ parameters, as we can solve for them given their respective (known) mean and standard deviation values,
\begin{multicols}{2}
  \noindent
  \begin{equation}
    \mu_{\text{unit uniform}} = 0.5, \sigma_{\text{unit uniform}} = \nicefrac{1}{\sqrt{12}}.
  \end{equation}
  \begin{equation}
    \mu_{\text{Dirac delta}} = \upsilon_{i}, \sigma_{\text{Dirac delta}} = 0.
  \end{equation}
\end{multicols}

We linearly interpolate  the mean $\mu$ and the standard deviation $\sigma$ parameters to obtain the intermediate Beta distribution parameterisations in between the two extremes,
\begin{multicols}{2}
  \noindent
  \begin{equation}
    \mu := \phi \upsilon_{i} + (1 - \phi) \mu_{\text{uniform}}
  \end{equation}
  \begin{equation}
    \sigma := (1 - \phi) \sigma_{\text{uniform}}.
  \end{equation}
\end{multicols}
Note that the standard deviation $\sigma$ will approach (but never reach) zero as of $\phi < 1$.

We then solve for $\alpha$ and $\beta$ using the analytical mean and standard deviation of Beta distributions, resulting in
\begin{multicols}{2}
  \noindent
  \begin{equation}
    \alpha = c_1 \beta
  \end{equation}
  \begin{equation}
    \beta = \frac{c_1 - c_2}{c_2 (c_1 + 1)},
  \end{equation}
\end{multicols}
where $c_1 = \frac{\mu}{1 - \mu}$ and $c_2 = \sigma^2 (c_1 + 1)^2$.

In the supplement we include an animation showing intermediate Beta distributions for various $\phi$ around a fixed point ($\upsilon_{i} = 0.75$).

\section{Background}
\label{sec:problem_detail}

In~\cite{zela2019understanding} it was shown that detrimental solutions, in particular solutions exhibiting an overly large number of skip connections, coincide with high validation loss curvatures.
In their work, they view these as problematic solutions within the solution set of the model.
They propose regularisation on the weight space and early stopping, which they show is helpful in avoiding reaching these solutions. 
\cite{chu2020fair} instead proposes a change to the model, where different operation edges between the same nodes are not mutually exclusive,
and they also propose a regularisation term pushing edge weights towards either zero or one. 
These alterations they show are beneficial for avoiding an over-reliance on skip connections, as well as reducing the approximation error resulting from the discretisation of the edge weights happening between the search and evaluation phase.
In addition, they made the solution set more expressive as of allowing multiple simultaneous operations between the same nodes of a cell.
In our work, we show that~\cite{chu2020fair} still suffers from another detrimental effect, 
similar to the one it was addressing, 
indicating that the issue has not yet been solved in full.
Similar to DARTS~\cite{liu2018darts} and RobustDARTS~\cite{zela2019understanding}, FairDARTS~\cite{chu2020fair} constructs the architecture from copies of a \emph{normal cell} and a \emph{reduction cell}.
What we show is that, using FairDARTS, 
the search is required to be stopped early to avoid reaching solutions that are detrimental to test performance during the evaluation phase or ultimately reaching invalid solutions post-discretisation.
Notably, the architecture - as described by its operation edge weights - changes very little from very early on in the search until it is stopped.
After the epoch it would have been stopped, 
the operation edge weights belonging to nodes in the normal cell all tend to zero.
Following discretisation of the edge weights, the normal cell no longer propagates activations through, making the architecture invalid.

We suggest that the cause of this problem is that the detrimental solutions correspond to local minima in the edge weights space, given typical initial positions in the neural network parameters space.
In particular, 
that as a consequence of the reduction cell operations relying on fewer parameters than normal cell operations, 
such solutions take up a large volume of the neural parameter space.

To see this, 
let us consider a detrimental solution $\{\bm{\alpha}, \bm{w}\}_{\text{detrimental}}$,
where all $\bm{\alpha}_{\text{normal}}$ elements are close to zero.
As will be confirmed in experiments, 
the neural network is sufficiently flexible to produce low loss solutions despite these elements being \emph{close} to zero.
Note that as long as activations can propagate through the normal cell, the reduction cell, being sufficiently expressive, can still represent low loss mappings.
Furthermore, 
for constellations where the operations in the normal cell have little to no effect on the loss, 
this directly translates into invariance to all of the associated $\bm{w}_{\text{normal}}$ neural network parameters.
In other words, such solutions are "large" in the sense that functionally equivalent solutions exist at all positions in the $\bm{w}_{\text{normal}}$ subspace.
We may think of this as an equivalent solution set. 

Secondly, consider a random initial set of neural network parameter values, $\bm{w}_{\text{initial}}$.
The "larger" an equivalent solution set is, the more likely it is that $\bm{w}_{\text{initial}}$ will end up inside or "close" to it.
In general, as well known and studied in the optimisation literature,
gradient-based local optimisation is subject to finding local which are not necessarily global minima.
In many applications, 
such as optimisation of neural network parameters alone, 
a local minimum might be "good enough".
However, in this application, if it is applied to $\bm{\alpha}_{\text{normal}}$, it may add a bias towards local solutions, being compatible edge weights with the initial values of the neural network parameters.

\section{Differentiable Neural Architecture Search}

\subsection{Architecture}

DARTS~\cite{liu2018darts}, 
as similar to prior works~\cite{zoph2018learning,real2019regularized,liu2018progressive},
searches for a \emph{cell} as the building block for the final architecture.
In the case of convolutional networks, the cell is stacked, and for recurrent networks, it is recursively connected.

The cell constitutes a directed acyclic graph of $N$ nodes.
Each node $x$ represents a latent representation
and each directed edge $(i, j)$ represents an operation $o_{i, j}$.
A node depends on all of its predecessors as
\begin{equation}
    x_j = \sum_{i < j} o_{i, j}(x_i).
\end{equation}
The cell is assumed to have two input nodes and a single output node.
In the case of convolutional networks, the input nodes are the outputs of the previous two layers,
and for recurrent cells, the input nodes represent the current step, and the state carried from the previous step. 
The cell output is obtained by a reduction operation (e.g. concatenation) to all the intermediate nodes.

Let $\mathcal{O}$ be the set of candidate operations (e.g., convolution, max pooling, skip connection) available for each edge $(i, j)$.
\cite{liu2018darts} proposed a relaxation over the discrete operation choice using softmax
\begin{equation}
    \bar{o}_{i, j}(x) = 
    \sum_{o \in \mathcal{O}}
    \frac{
    \text{exp}(\alpha_{o_i,j})
    }{
    \sum_{o' \in \mathcal{O}} \text{exp}(\alpha_{o'_i,j})
    }
    o(x),
    \label{eq:softmax}
\end{equation}
where the operation weights for a pair of nodes $(i, j)$ are parameterised by a vector $\bm{\alpha}_{i, j}$ of
dimension $|\mathcal{O}|$.
Importantly, this makes the search space continuous and allows gradient-based optimisation methods.

FairDARTS~\cite{chu2020fair}, building upon~\cite{liu2018darts}, proposed replacing~Eq.~\ref{eq:softmax} with
\begin{equation}
    \bar{o}_{i, j}(x) = 
    \sum_{o \in \mathcal{O}}
    \sigma(\alpha_{o_i,j})
    o(x)
    \label{eq:sigmoid}
\end{equation}
where $\sigma$ is the sigmoid function. 
This allows multiple operations per edge to be chosen simultaneously.
If no operations are active for a given edge, this constitutes the zero operation~\cite{zela2019understanding}.

For the case of convolutional neural networks, 
on which we will focus in this paper, 
both DARTS and FairDARTS searches for a normal
cell and a reduction cell to build up the final architecture.
The reduction cell, 
in contrast to the 'normal' cell, 
reduces the number of activation maps (or channels) out from the cell.

\subsection{Search}
\label{sec:search}

Let $\bm{\alpha}$ be the concatenated vector of all operation edge weights representing the architecture,
and $\bm{w}$ be the concatenated neural network parameters associated with all operations.
The $\bm{\alpha}$ vector contains the operation edge weights associated with both the normal cell and the reduction cell that are being searched for, i.e. $\bm{\alpha} = \{\bm{\alpha}_{\text{normal}}, \bm{\alpha}_{\text{reduction}}\}$, and the same applies to the weight parameters, $\bm{w} = \{\bm{w}_{\text{normal}}, \bm{w}_{\text{reduction}}\}$.

The architecture search problem was in~\cite{liu2018darts} stated as the bi-level optimisation problem
\begin{mini!}|l|[2]
{\bm{\alpha}}
{
\mathcal{L}_{\text{val}}(\bm{\alpha}, \bm{w}^{\ast})
}
{}{
\label{eq:bilevel_1}
}
\addConstraint{
\bm{w}^{\ast} = \argmin_{\bm{w}} \mathcal{L}_{\text{train}}(\bm{\alpha}, \bm{w})
},
\label{eq:bilevel_2}
\end{mini!}
where $\mathcal{L}_{\text{val}}$ and $\mathcal{L}_{\text{train}}$ are the validation loss and training loss, respectively.

The proposed optimisation procedure in~\cite{liu2018darts} is to approximate the gradient as
\begin{equation}
    \nabla_{\bm{\alpha}} \mathcal{L}_{\text{val}}(\bm{\alpha}, \bm{w}^{\ast}) 
    \approx 
    \nabla_{\bm{\alpha}} \mathcal{L}_{\text{val}}(
    \bm{\alpha}, 
    \bm{w} - 
    \xi \nabla_{\bm{w}} \mathcal{L}_{\text{train}}(\bm{\alpha}, \bm{w})
    )
\end{equation}
and perform gradient-based local optimisation,
alternating between taking a step in the optimisation problem of $\argmin_{\bm{\alpha}} \mathcal{L}_{\text{val}}$ and of $\argmin_{\bm{w}} \mathcal{L}_{\text{train}}$.
$\bm{w}$ are the current weights and $\xi$ is the learning rate for a step in the inner optimisation problem (Eq.~\ref{eq:bilevel_2}).
This can be described as, at iteration $t$, take steps using the gradients defined at
\begin{equation}
        \nabla_{\bm{\alpha}} \mathcal{L}_{\text{val}}(\bm{\alpha}_{t}, \bm{w}_{t}),
\end{equation}
followed by 
\begin{equation}
        \nabla_{\bm{w}} \mathcal{L}_{\text{train}}(\bm{\alpha}_{t+1}, \bm{w}_{t}),
\end{equation}
where $\bm{\alpha}_{t}$ and $\bm{w}_{t}$ is the position of respective parameter at the beginning of the iteration, and $\bm{\alpha}_{t+1}$ and $\bm{w}_{t+1}$ the updated positions, respectively.

\section{Global Optimisation Scheme}
\label{sec:optimization_scheme}

\begin{algorithm}[t]
  \SetAlgoLined
  \SetNoFillComment
  \KwIn{Function $f$ defined over $\mathcal{X}$, initial $\bm{x}_{\text{best}}, y_{\text{best}}$, local\_step, global\_step}
  \KwOut{$\bm{x}_{\text{best}}, y_{\text{best}}$}
  reset schedule;\\
  \While{budget remaining}{
    take\_global\_step = schedule.current()\;
    \uIf{take\_global\_step}{
    $\bm{x}_{\text{best}}$, $y_{\text{best}}$ = global\_step($\bm{x}_{\text{best}}$, $y_{\text{best}}$)
    \;
    $\bm{x}_{\text{current}}, y_{\text{current}} = \bm{x}_{\text{best}}, y_{\text{best}}$\;
    }
    \Else{
    $\bm{x}_{\text{current}}$, $y_{\text{current}}$ = local\_step($\bm{x}_{\text{current}}$, $y_{\text{current}}$)\;
    \If{$y_{\text{current}} < y_{\text{best}}$}{
     $\bm{x}_{\text{best}}, y_{\text{best}} = \bm{x}_{\text{current}}, y_{\text{current}}$\;
    }
    }
    schedule.step($y_{\text{best}}$)\;
  }
  return $\bm{x}_{\text{best}}, y_{\text{best}}$\;
  \caption{Local optimisation with global optimisation backtracking}
  \label{algo:hybrid}
\end{algorithm}

In Section~\ref{sec:problem_detail} we hypothesised that the usage of local search for the $\bm{\alpha}$ weights adds bias towards solutions
compatible with $\bm{w}$ solutions that are closer to the initial position.
We will now describe a simple hybrid scheme, which makes the search for the $\bm{\alpha}$ weights "more global" in that it is less subject to the local curvature of the loss surface.
We will later evaluate this scheme empirically, contrasting it to the previous, fully local search.

The $\bm{\alpha}$ parameter, 
being the collection of operation edge weights representing the architecture, 
is typically vastly different than the neural network parameters $\bm{w}$ 
in dimensionality.
$\bm{\alpha}$ is, for the search spaced addressed, $196$-dimensional, while $\bm{w}$ has millions of parameters.
In FairDARTS and other DARTS variants, both parameters are optimised using gradient-based local optimisation, with alternating steps as described in Section~\ref{sec:search}.
However, the moderate dimensionality of the $\bm{\alpha}$ parameter makes it practically feasible to apply global optimisation techniques to optimise it.
Specifically, we will make use of the idea of coordinate descent, where one coordinate is optimised at a time,
as well as annealed sampling.
In this section, we will describe a simple hybrid between local and global optimisation,
which we later show performs well empirically.

We will first outline the general algorithm of the hybrid approach in Algorithm~\ref{algo:doubly_coord},
in turn, parameterised by functions responsible for taking a "local" step and "global" step, respectively.
At this abstraction level, we only distinguish between a global and local step, by if after taking the step, the "current position" is the same as the "best observed" position so far, in terms of smallest loss.
For brevity, 
we leave out that the $global\_step$ function considers all observations so far,
without loss of generality.
The remaining components are specified in Section~\ref{sec:globalopt} and Section~\ref{sec:beta}.

\end{document}